%% file: main.tex
\documentclass{article}

    \usepackage[nonatbib,preprint]{nips_2018}

    \usepackage[utf8]{inputenc} 
    \usepackage[T1]{fontenc}    
    \usepackage{url}            
    \usepackage{booktabs}       
    \usepackage{amsfonts}       
    \usepackage{nicefrac}       
    \usepackage{microtype}      

    \usepackage{color}
    \usepackage{epsfig}
    \usepackage{graphicx}
    \usepackage{amssymb}
    \usepackage{paralist}
    \usepackage[toc,page]{appendix}
    \usepackage{wrapfig}
    \usepackage{subcaption}
    \usepackage[font=footnotesize,labelfont=footnotesize]{caption}

    \usepackage{amssymb}
    \usepackage{pifont}
    \newcommand{\cmark}{\ding{51}}%
    \newcommand{\xmark}{\ding{55}}%

    \usepackage[utf8]{inputenc} 
    \usepackage[T1]{fontenc}    
    \usepackage{url}            
    \usepackage{booktabs}       
    \usepackage{amsfonts}       
    \usepackage{nicefrac}       
    \usepackage{microtype}      
    \usepackage[pagebackref=true,breaklinks=true,colorlinks,bookmarks=false,citecolor=blue]{hyperref} 
    \usepackage{mysymbols}
    \usepackage{enumitem}

    \input{space_saver}
    \title{EvalAI: \\
    Towards Better Evaluation Systems for AI Agents}
    
    \author{
        Deshraj Yadav$^1$
        \and
        \textbf{Rishabh Jain}$^1$
        \and
        \textbf{Harsh Agrawal}$^1$
        \and
        \textbf{Prithvijit Chattopadhyay}$^1$ \\
        \and
        \textbf{Taranjeet Singh}$^2$
        \and
        \textbf{Akash Jain}$^3$
        \and
        \textbf{Shiv Baran Singh}$^4$ \\
        \and
        \textbf{Stefan Lee}$^1$
        \and
        \textbf{Dhruv Batra}$^{1}$ \\
        $^1$Georgia Institute of Technology \quad $^2$Paytm \quad $^3$ Zomato \quad $^4$Cyware \\ \and
        \texttt{\{deshraj,rishabhjain,harsh.agrawal,prithvijit3,steflee,dbatra\}@gatech.edu}\\ 
        \and
        \texttt{\{reachtotj,akashjain993,spyshiv\}@gmail.com} \and
    }

\begin{document}
    
    \maketitle
    
    \begin{abstract}{
    We introduce \href{https://evalai.cloudcv.org}{EvalAI}, an open source platform for evaluating and comparing machine learning (ML) and artificial intelligence algorithms (AI) at scale. 
    EvalAI is built to provide a scalable solution to the research community to fulfill the critical need of evaluating machine learning models and agents acting in an environment against annotations or with a human-in-the-loop.
    This will help researchers, students, and data scientists to create, collaborate, and participate in 
    AI challenges organized around the globe. By simplifying and standardizing the process of benchmarking these models, EvalAI seeks to lower the barrier to entry for participating in the global scientific effort to push the frontiers of machine learning and artificial intelligence, thereby increasing the rate of measurable progress in this domain.
    \\ Our code is available \href{https://github.com/Cloud-CV/EvalAI}{here}}.
    \end{abstract}

    \input{sections/intro.tex}
    \input{sections/related_work.tex}
    \input{sections/solution.tex}

\input{sections/infrastructure.tex}
    \input{sections/use_cases.tex}
    \input{sections/case_studies.tex}
    \input{sections/conclusion.tex}

    {
    \bibliographystyle{plain}
    \bibliography{main}
    }

    \end{document}

%% file: space_saver.tex
\belowdisplayskip \abovedisplayskip

\newenvironment{packed_item}{
\begin{itemize}[leftmargin=*]
  \setlength{\itemsep}{1pt}
  \setlength{\parskip}{0pt}
  \setlength{\parsep}{0pt}
}{\end{itemize}}

\newlength{\sectionReduceTop}
\newlength{\sectionReduceBot}
\newlength{\subsectionReduceTop}
\newlength{\subsectionReduceBot}
\newlength{\abstractReduceTop}
\newlength{\abstractReduceBot}
\newlength{\captionReduceTop}
\newlength{\captionReduceBot}
\newlength{\subsubsectionReduceTop}
\newlength{\subsubsectionReduceBot}

\newlength{\eqnReduceTop}
\newlength{\eqnReduceBot}

\newlength{\horSkip}
\newlength{\verSkip}

\newlength{\figureHeight}

%% file: sections/intro.tex
\section{Introduction}
\label{sec:intro}
Time and again across different scientific and engineering fields, the formulation and creation of the right question, task, and dataset to study a problem has coalesced fields around particular challenges --  driving scientific progress. Likewise, progress on important problems in the fields of Computer Vision (CV) and Artificial Intelligence (AI) has been driven by the introduction of bold new tasks together with the curation of large, realistic datasets \cite{krizhevsky_nips12,{VQA}}.

 Not only do these tasks and datasets establish new problems and provide data necessary to address them, but importantly they also establish reliable benchmarks where proposed solutions and hypothesis can be tested which is an essential part of the scientific process. In recent years, the development of centralized evaluation platforms have lowered the barrier to compete and share results on these problems. As a result, a thriving community of data scientists and researchers has grown around these tasks, increasing the pace of progress and technical dissemination.

\begin{figure}[ht!]
\centering
\hspace*{-1.5cm}\includegraphics[width=1.13\columnwidth]{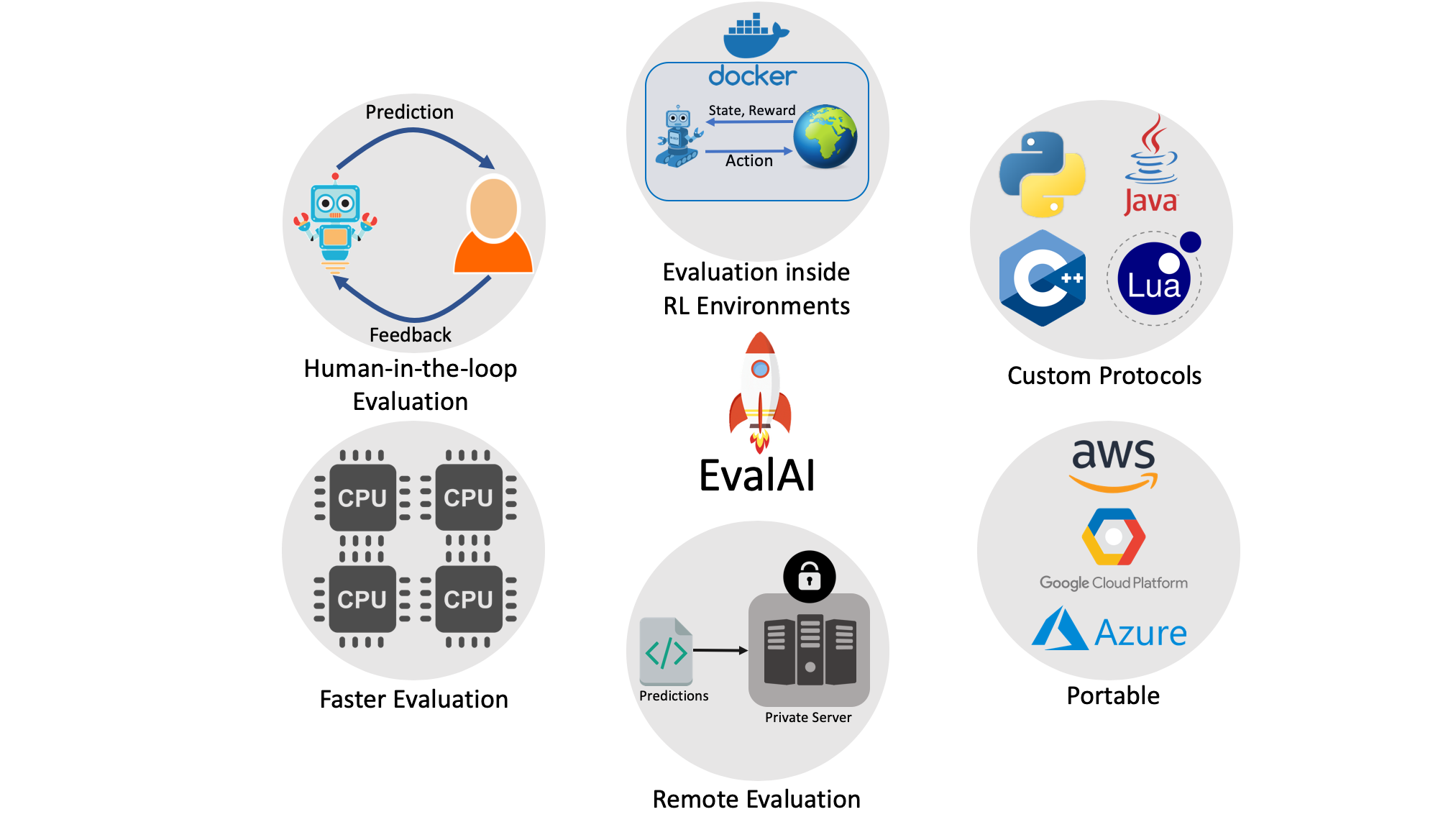}
\caption{EvalAI is a new evaluation platform with the overarching goal of providing the right tools, infrastructure and framework to setup exhaustive evaluation protocols for both traditional static evaluation tasks as well as those in dynamic environments hosting multiple agents and/or humans.}
\label{fig:teaser}
\end{figure}

\par Historically, 
the community has focused on 
traditional AI tasks such as image classification, scene recognition, and sentence parsing that follow a standard \emph{input-output} paradigm for which models can be evaluated in isolation using simple automatic metrics like accuracy, precision or recall. But with the success of deep learning techniques on a wide variety of tasks and the proliferation of `smart' applications, there is an imminent need to evaluate AI systems in the context of human collaborators and not just in isolation.
This is especially true as AI systems become more commonplace and we find ourselves interacting with AI agents on a daily basis. For instance, people frequently interact with virtual assistants like Alexa, Siri, or Google Assistant to get answers to their questions, to book appointments at a restaurant, or to reply to emails and messages automatically. Another such example is the use of AI for recognizing content in images, helping visually impaired users interpret the surrounding scene. To this end, the AI community has introduced several challenging high-level AI tasks
ranging from question-answering about multiple modalities (short articles ~\cite{Rajpurkar2016SQuAD10}, images~\cite{antol_iccv15}, videos~\cite{Tapaswi2016MovieQAUS}) to goal oriented dialog ~\cite{Bordes2016LearningEG} to agents acting in simulated environments to complete task-specific goals~\cite{embodiedqa, kolve2017ai2, zhu2017target}.  

\par As AI improves and takes on these increasingly difficult, high-level tasks that are poorly described by an \emph{input-output} paradigm, robust evaluation faces a number of challenges. For instance, generating a natural language description for an image, having a conversation with a human, or generating aesthetically pleasing images cannot be evaluated accurately using automatic metrics as performance on these metrics do not correlate well with human-judgment in practice~\cite{visdial_eval}. Such tasks naturally require human-in-the-loop evaluation by connecting the AI system with a human workforce such as Amazon Mechanical Turk (AMT)~\cite{amt} to quantify performance in a setup which is closest to the one in which they may be eventually deployed. Moreover, human-AI interaction also reveal interesting insights into the true capabilities of machine learning models. For instance,~\cite{visdial_eval} connected human users with AI agents trained to answer questions about images in a 20-questions style image guessing game and then measured the performance of the human-AI team. The authors observed from the experiments that surprisingly, performance gained through AI-AI self-play does not seem to generalize to human-AI teams. These sort of useful insights are increasingly becoming important as more and more of these models reach consumers.

\par Furthermore, the rise of reinforcement learning based problems in which an agent must interact with an environment introduces additional challenges for benchmarking. In contrast to the supervised learning setting where performance is measured by evaluating on a static test set, it is less straightforward to measure generalization performance of these agents in context of the interactions with the environment.
Evaluating these agents involves running the associated
code on a collection of unseen environments that constitutes a hidden test set for such a scenario.
\par To address the aforementioned problems, we introduce a new evaluation platform for AI tasks called EvalAI. It is an extensible open-source platform that fulfills the critical need in the community for (1) human-in-the-loop evaluation of machine learning models and (2) the ability to run user's code in a dynamic environment 
to support the evaluation of interactive agents.  
We have also addressed several limitations of existing platforms by supporting (3) custom evaluation pipelines compatible with
any programming language, (4) arbitrary number of challenge phases and dataset splits, and (5) remote evaluation on private worker pool. By providing the functionality to connect agents, environments, and human evaluators in one single platform, EvalAI enables novel research directions to be explored quickly and at scale. 

\section{Desiderata}
Having outlined the need for an evaluation platform that can properly benchmark increasingly complex AI tasks, in this section we explicitly specify the following \emph{requirements} that a modern evaluation tool should satisfy. 

\par \noindent
\textbf{Human-in-the-loop evaluation of agents.} As discussed in the Sec.~\ref{sec:intro}, the AI community has introduced increasingly bold tasks 
(goal-oriented dialog, question-answering, GuessWhich, image generation, etc.) 
some of which require pairing the AI agent with a human 
to accurately evaluate and benchmark different approaches against each other. A modern evaluation platform should provide a unified framework for benchmarking in scenarios in which agents are not acting in isolation, but are  rather interacting with other agents or humans at test-time.
 
\par \noindent
\textbf{Environments, not datasets.} As the community becomes more ambitious in the problems they are trying to solve, we have noticed a shift from static datasets to dynamic environments. Now instead of evaluating a model on a single task, agents are deployed in new unseen environments inside a simulation to check for generalization to novel, unseen scenarios~\cite{kolve2017ai2,embodiedqa}.
As such, modern evaluation platforms need to be capable of running ``submitted'' agents within these environments -- a significant departure from the standard evaluation paradigm of computing automatic metrics on a set of submitted predictions.

\par \noindent
\textbf{Extensibility.} Different tasks require different evaluation protocols. An evaluation platform needs to support an arbitrary number of phases and dataset splits to cater to the evaluation scenarios which often use multiple dataset splits, each serving a different purpose. For instance, COCO Challenge~\cite{lin2014microsoft}, VQA~\cite{VQA}, and Visual Dialog~\cite{visdial} 
all use multiple splits such as test-dev for validation, test-std for reporting results in a paper and a separate test-challenge split for announcing the winners of a challenge that may be centered around the task. 

%% file: sections/related_work.tex
\section{Related Work}
\label{sec:related_work}
Here we survey some of the existing evaluation platforms in light of the requirements highlighted in the previous section. Additionally, for reader's convenience, we summarize the features offered by EvalAI via head-to-head comparison with the existing platforms in Table.~\ref{tab:platform_compare}.

\begin{table}[t]
\centering
\resizebox{\textwidth}{!}{%
\renewcommand*{\arraystretch}{1.2}
\begin{tabular}{lcccccccc}
\toprule
\textbf{Features} & \textbf{OpenML} & \textbf{Topcoder} & \textbf{Kaggle} & \textbf{CrowdAI} & \textbf{ParlAI} & \textbf{CodaLab} & \textbf{EvalAI} \\ \midrule
AI Challenge Hosting &  \xmark & \cmark & \cmark & \cmark & \xmark & \cmark & \cmark \\ 
Custom metrics & \xmark &  \xmark & \xmark & \cmark & \cmark& \cmark & \cmark \\ 
Multiple phases/splits &  \xmark & \xmark & \xmark & \cmark & \xmark & \cmark & \cmark \\ 
Open Source &  \cmark & \xmark & \xmark & \cmark & \cmark & \cmark & \cmark \\ 
Remote Evaluation &  \xmark & \xmark & \xmark & \xmark & \cmark & \cmark & \cmark \\ 
Human Evaluation &  \xmark & \xmark & \xmark & \xmark & \cmark & \xmark & \cmark \\ 
Environments & \xmark & \xmark & \xmark & \cmark & \xmark & \xmark & \cmark \\\bottomrule
\end{tabular}
}
\vspace{10pt}
\caption{Head-to-head comparison of capabilities between existing platforms and EvalAI} \label{tab:platform_compare}
\vspace{-10pt}
\end{table}

\par \textbf{Kaggle} \cite{Kaggle} is one of the most popular platforms for hosting data-science and machine learning competitions. 
It allows users to share their approach with other data scientists through a cloud-based workbench that is similar to IPython notebooks in terms of functionality. Despite it's
popularity, Kaggle has several limitations. Firstly, being a closed-source platform limits 
Kaggle from supporting complex AI tasks that require customized metrics other than the usual ones
available through the platform. Secondly, it does not allow multiple challenge phases -- a common practice in popular challenges like VQA, Visual Dialog, COCO Caption Challenge. 
Lastly, Kaggle does not allow hosting a challenge with a private test splits or workers, thereby limiting the platforms usability in scenarios where organizers cannot share the
test-set publicly.

\par \textbf{CodaLab} \cite{CodaLab} is another open-source alternative to Kaggle providing an ecosystem for conducting computational research in a more efficient, reproducible, and collaborative manner. There are two aspects of CodaLab -- worksheets and competitions. 
Worksheets enable users to capture complex research pipelines in a reproducible way, creating ``executable papers'' in the process. By archiving the code, data and the results of an experiment, the users can precisely capture the research pipeline in an immutable way. Additionally, it enables a researcher to present these pipelines in a comprehensible way using worksheets or notebooks written in a custom markdown language. 
CodaLab Competitions provide an evaluation platform on which supports hosting competitions and benchmarking through a
public leaderboard.
While CodaLab Competitions is very similar to EvalAI and addresses some of the limitations of Kaggle in terms of functionality, it does not support evaluating interactive agents in
different environments with or without humans in the loop.
As the community introduces more complex tasks in which evaluation requires running an agent inside a simulation or pairing an agent with a human workforce for evaluation, a highly customizable backend like ours connected with existing platforms like Amazon Mechanical Turk become extremely important.

\par \textbf{OpenML} \cite{OpenML2013} is an online platform where researchers can automatically log and share machine learning data sets, code, and experiments. As a system, OpenML allows people to organize their experiments online, and build directly on top of the work of others. By readily integrating the online platform with several machine learning tools, large-scale collaboration in real-time is enabled,
allowing researchers to share their very latest results while keeping track of their impact and use. While the focus of OpenML is on experiments and datasets, EvalAI focuses more on the end result - models, their predictions and subsequent evaluation. OpenML and EvalAI are complementary to each other and will be useful to the user at different stages of the research.

\par \textbf{ParlAI}~\cite{miller2017parlai} is a recently introduced open-source platform for dialog research implemented in Python. It serves as a unified platform for sharing, training and evaluating models for several dialog tasks. Additionally, ParlAI also supports integration with Amazon Mechanical Turk -- to collect dialog data and support human-evaluation. Several popular dialog datasets and tasks are supported off-the-shelf in ParlAI. Note that unlike EvalAI, ParlAI supports only evaluation for dialog models not for any AI task in general. Also, unlike EvalAI -- which supports evaluation across multiple-phases and splits to truly test the generalization and robustness of the proposed model, ParlAI only supports evaluation on one test split, as is the norm with most of the existing dialog datasets.

\par Additionally, reinforcement learning (RL) algorithms also require
strong evaluation and good benchmarks. A variety of benchmarks have been released, such as the Arcade Learning Environment (ALE) \cite{DBLP:journals/corr/abs-1207-4708}, which exposed a collection of Atari 2600 games as reinforcement learning problems, and recently the RLLab benchmark for continuous control \cite{DBLP:journals/corr/DuanCHSA16}. 
More recently, \textbf{OpenAI} \cite{BrockmanCPSSTZ16} gym was released as a toolkit to develop and compare RL algorithms on a variety of environments and tasks -- ranging from walking to playing pong or pinball. The gym library provides a flexible environment in which agents can be evaluated using existing machine learning frameworks, such as TensorFlow or PyTorch. OpenAI gym has a similar underlying philosophy of encouraging easy accessibility and reproducibility by not restricting to any particular framework. Additionally, environments are versioned in a way
to ensure meaningful and reproducible results as the software is updated.

%% file: sections/solution.tex
\section{EvalAI: Key Features}
\label{sec:key_features}
As discussed in the previous sections, ensuring algorithms are compared fairly in a standard way is a difficult and ultimately distracting task for AI researchers.
Establishing fair comparison requires rectifying minor differences in algorithm inputs,
implementing complex evaluation metrics,
and often ensuring the correct usage of non-standard dataset splits. In the following sub-sections, we describe the key features that 
address the aforementioned problems.
\subsection{Custom Evaluation Protocol}

EvalAI is highly customizable since it allows creation of an arbitrary number of evaluation phases and dataset splits, compatibility using any programming language, and organizing results in both public and private leaderboards. All these services are available through an intuitive web-platform and comprehensive REST APIs. 

\subsection{Human-in-the-loop Evaluation}

\par While standard computer vision tasks such as image classification~\cite{simonyan_iclr15,he2016deep}, semantic or instance segmentation~\cite{long2015fully,he2017mask}, object detection~\cite{he2017mask,redmon2016you} are easy to evaluate using the automatic metrics, it is notoriously difficult to evaluate tasks for which automated metrics correlate poorly with human judgement -- for instance,
natural language generation tasks such as image captioning ~\cite{dai2017towards,li2018generating}, visual dialog \cite{visdial, visdial_rl} or image generation tasks ~\cite{goodfellow_nips14}. 
Developing measures which correlate well with human judgment remains an open area of research. Automatic evaluation of models for these kind of tasks is further complicated by the huge set of possibly `correct' or `plausible' responses and the relatively sparse set of ground truth annotations, even for large-scale datasets.

\par Given these difficulties and the interactive nature of tasks, it is clear that the most appropriate way to evaluate these kind of tasks is with a human in the loop, i.e. a Visual Turing Test \cite{geman2015visual}! Unfortunately, large-scale human-in-the-loop evaluation is still limited by financial and infrastructural challenges that must be overcome by each interested research group independently. Consequently, human evaluations are rarely performed and experimental settings vary widely, limiting the usefulness of these benchmarking studies in human-AI collaborative settings.

\par We propose to fill this critical need in the community by providing the capability of human-in-the-loop evaluation. To this end, we have developed the infrastructure to pair Amazon Mechanical Turk (AMT) users in real-time with artificial agents -- specifically visual dialog as an example.

\subsubsection{Challenges}
Bulding a framework to support human-in-the-loop evaluation comes with it's own set of challenges:
\begin{packed_item}
\item \textbf{Instructions}: Since the workers do not know their roles before starting a study catered towards evluating such tasks, they need detailed instructions and a list of Do's and Dont's for the task. Each challenge might have different instructions and therefore we provide challenge organizers the flexibility to provide us with the required instructions in their own HTML templates.

\item \textbf{Worker pool}: We need to ensure that we have a pool of good quality workers who have prior experience in doing certain tasks and have a history of high acceptance rate(s). We allow organizers to provide us with a list of whitelisted and blocked workers. Additionally, they can also provide a qualification test which the workers need to pass to participate in the evaluation tasks.

\item \textbf{Uninterrupted back-and-forth communication}: Certain tasks like evaluating dialog agents need uninterrupted back-and-forth communication between agents and workers. However, this is not always possible since turkers might disconnect or close a HIT before finishing it. We do extensive book-keeping to ensure that incompleted HITS are re-evaluated and turkers can reconnect with the same agent if the connection was interrupted only temporarily. 
\item \textbf{Gathering results}: We provide a flexible JSON based schema and APIs to fetch the results from the evaluation tasks once they are completed. These results are automatically updated on the leaderboard for each submission.
\end{packed_item}

\vspace{-5pt}
\subsection{Remote Evaluation}
\vspace{-5pt}
\begin{figure}[h]
\centering
\includegraphics[width=0.8\columnwidth]{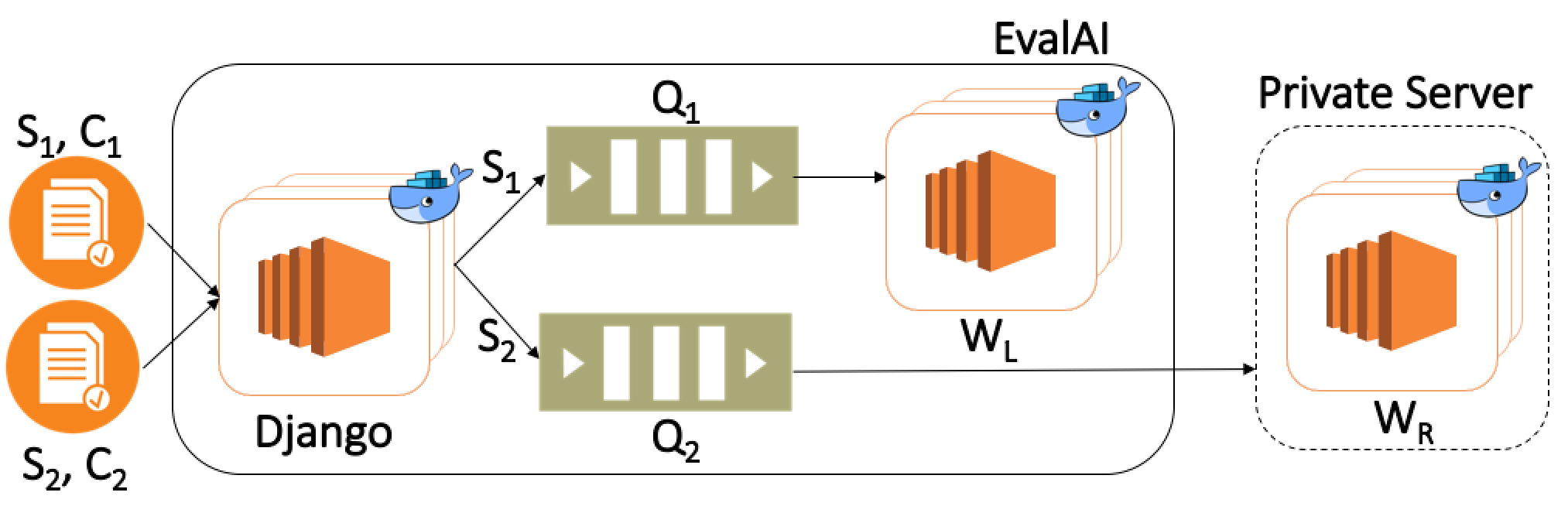}
\caption{Remote Evaluation Pipeline: Challenge $C_1$ and $C_2$ are hosted on EvalAI but evaluation for $C_2$ happens on an evauation worker that is running on a private server which is outside EvalAI Virtual Private Cloud (VPC). For two submissions $S_1$ and $S_2$ made to challenges $C_1$ and $C_2$ respectively, submission $S_1$ will be evaluated on $W_L$ which is running on EvalAI whereas $S_2$ will run on $W_R$ which is a remote machine.}
\label{fig:remote_eval}
\end{figure}

Certain large-scale challenges need special compute capabilities for evaluation. For instance, running an agent based on some deep reinforcement learning model in a dynamic environment will require powerful clusters with GPUs. If the challenge needs extra computational power, challenge organizers can easily add their own cluster of worker nodes to process participant submissions while we take care of hosting the challenge, handling user submissions and the maintaining the leaderboard. Our remote evaluation pipeline (shown in \figref{fig:remote_eval}) decouples the worker nodes from the web servers through via message-queues. On submission, all related metadata is relayed to an external pool of workers through dedicated message queues.

%% file: sections/infrastructure.tex
\section{System Architecture}
\label{sec:system_architecture}

\begin{figure*}
\centering
\includegraphics[width=1\textwidth]{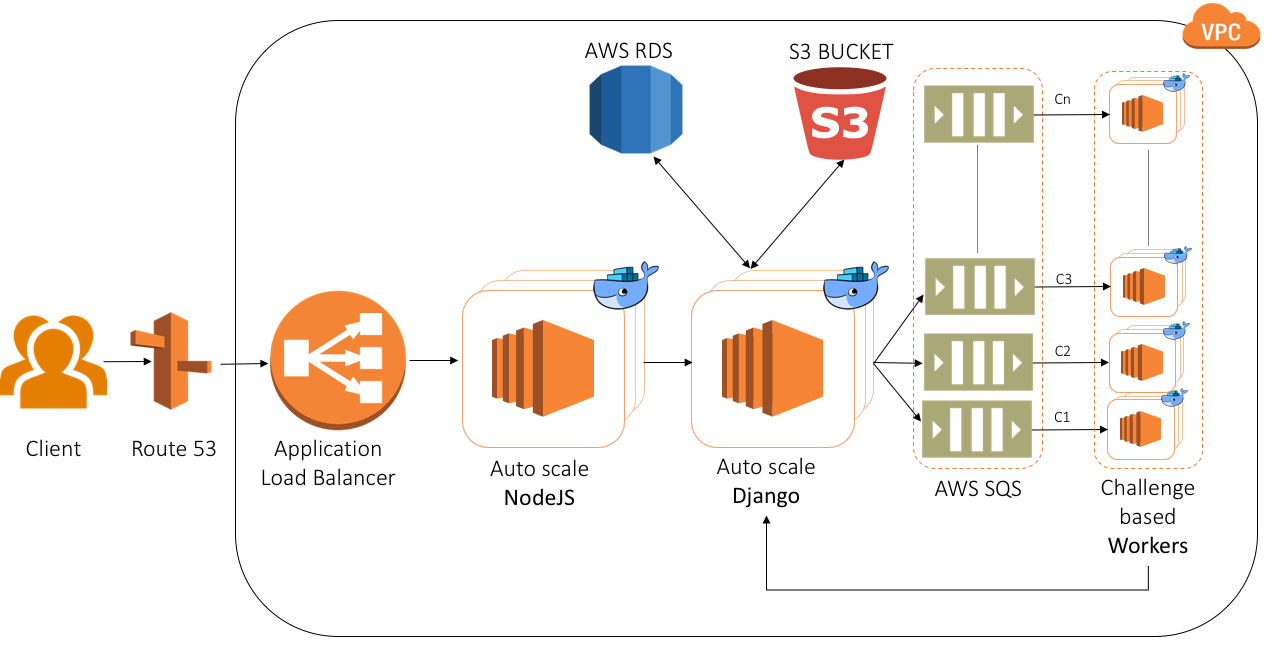}
\caption{System architecture of EvalAI}
\label{fig:architecture}
\vspace{-10pt}
\end{figure*}

The architectural back-end of our system (\figref{fig:architecture}) was designed with keeping in mind scalability and portability of such a system from the very inception of the idea. 
Most of the components rely heavily on open-source technologies -- Docker, Django, Node.js, and PostgreSQL. We also rely on certain proprietary services but at the same time we ensure that the protocol is consistent with other open-source alternatives to the best possible extent. The configurations to setup proprietary services are also available through our open-source code. In the following sub-sections, we describe in detail the key pieces of our platform.

\par \noindent
\textbf{Orchestration - }
We rely heavily on Docker \cite{docker} containers to run our infrastructure. Additionally, we also deploy all our containers on Amazon Elastic Container Service (ECS) \cite{ecs} which auto-scales the cluster to meet the computational demands of the platform leading to high operational efficiency.

\par \noindent \textbf{Web Servers - }
 EvalAI uses Django \cite{django} which is a Python based MVC framework that powers our backend. It is responsible for accessing and modifying the database using APIs, and submitting the evaluation requests into a queue. It also exposes certain APIs to serve data and fetch results from Amazon Mechanical Turk \cite{amt} during human evaluation. Through these APIs, agents and workers on AMT communicate with each other using JSON blobs. By structuring the communication protocol to JSONs, the challenge organizers are enabled to customize it to support any kind of interaction.

\par \noindent \textbf{Message Queue - }
The message queue is responsible for routing a user's submission to the appropriate worker pool based on the unique routing key associated with each challenge. For our message broker, we chose Amazon Simple Queue Service (SQS) \cite{sqs}. By using SQS, we do not have to worry about consistency and reliability of the queue. An added bonus of using SQS is that it works seamlessly with other AWS services we use. 

\par \noindent \textbf{Worker Nodes - }
For every challenge, there is a different pool of worker nodes dedicated to evaluating submissions specific to the challenge. We spawn worker nodes as docker containers running inside Elastic Container Service (ECS) \cite{ecs} which results in multiple advantages. First, worker nodes are isolated such that the dependencies for one challenge don't clash with dependencies of other challenges. Second, pool of worker nodes specific to the challenge can independently scale based on the demands of the challenge. We also worked closely with challenge organizers to optimize their code to leverage the full computational capacity of the worker. For instance, we \textbf{warm-up the worker nodes at start-up} by importing the challenge code and pre-loading the dataset in memory. We also split the dataset into small chunks that are simultaneously evaluated on multiple cores. These simple tricks result in faster evaluation and reduces the evaluation time by an order of magnitude in some cases. Refer to  Section \ref{sec:case_study} for details on speed-up for the VQA Challenge.

%% file: sections/use_cases.tex
\section{Lifecycle of a Challenge}
\label{sec:lifecycle}
We now describe the life-cycle of a challenge 
starting from
creating a challenge, submitting entries to the challenge and 
finally
evaluating the submissions. This process will 
also elaborate 
how different components of the platform communicate with each other.

\subsection{Challenge Creation}
There are two ways to create a challenge on our platform. For challenges like image classification, detection which require simple evaluation metrics (such as precision, recall, accuracy), a user can follow a sequence of prompts on a user-interface to create a challenge. For more complex scenarios which require custom evaluation metrics, multiple dataset splits and phases, users are recommended to create a competition bundle which specifies the challenge configuration, evaluation code, and information about the said data-splits. 
The associated configuration file provides enough flexibility to 
configure different phases of the competition, define 
number of splits for the dataset and specify custom evaluation scripts arbitrarily.

\subsection{Submission}
\begin{figure}[h]
\centering
\includegraphics[width=0.75\columnwidth]{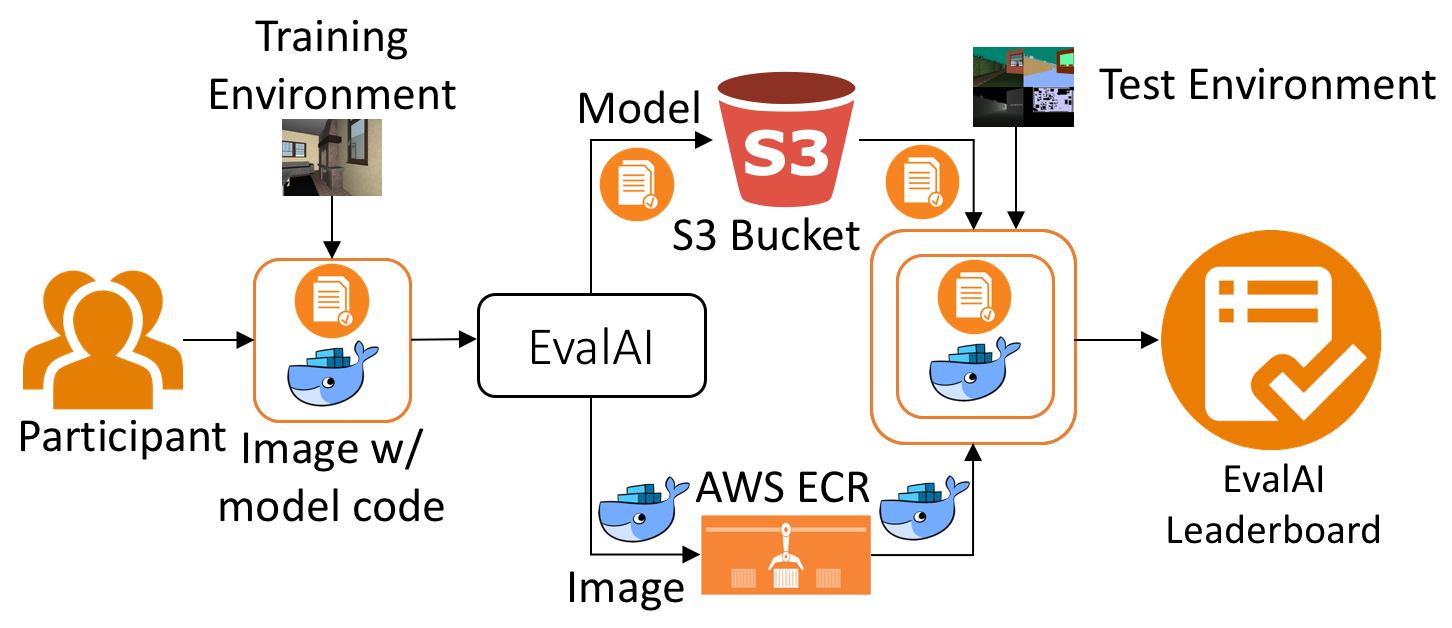}
\caption{EvalAI lets participants submit code for their agent which are eventually evaluated in dynamic environments on the evaluation server. The pipeline involves participants uploading the model snapshot and the code as docker image. Model snapshots are stored in Amazon S3 while the docker images are stored in Amazon Elastic Container Registry (ECR). During evaluation, the worker fetches the image, test environment and the model snapshot and spins up a new container to perform evaluation on this model. The results are then sent over to the leaderboard through a message queue.}
\label{fig:agent submission}
\end{figure}

EvalAI supports both submitting the model predictions and the model itself for evaluation. Traditional challenges require user to submit their model predictions on a static test set provided by the challenge organizers. On submission, these predictions are handed over to challenge specific workers that compare the predictions against corresponding ground-truth using the custom evaluation script provided by the challenge organizers. As we move towards developing intelligent agents for tasks situated in active environments instead of static datasets, where agents take actions to change the state of the world around them, it is imperative that we build new tools to accurately benchmark agents in environments. In this regard, we have developed an evaluation framework (shown in \figref{fig:agent submission})where participants upload Docker images with their pretrained models on Elastic Container Registry (ECR) and Amazon S3 respectively, which is then attached and run against test environments and evaluation metrics provided by the challenge organizer. At the time of evaluation, the instantiated worker fetches the image from ECR, assets and configuration for test-environment, model snapshot from Amazon S3 and spins up a new container to perform evaluation. Once the evaluation is complete, the results are sent over to the leaderboard using the message queue described in Section \ref{sec:system_architecture}.

\subsection{Evaluation}
We allow organizers to provide an implementation of their metric and is subsequently used to evaluate all submissions ensuring consistency in evaluation protocols. The by-product of containerizing the evaluation for different challenges in docker containers is that it allows us to package fairly complex pipelines, with all it's dependencies in an isolated environment. For human-in-the-loop evaluation, the evaluation code first loads up the worker and launches a new HIT on Amazon Mechanical Turk. Once the worker accepts the HIT, the worker is paired with the agent running inside a docker image. Based on the instruction given, the worker will interact with the agent and evaluate it according to certain criteria. This interaction data and the final rating given by the worker is stored by EvalAI which is eventually reflected on the leaderboard. EvalAI takes care of managing a persistent connection between the agent and the worker, error handling, retrying , storing the interaction data corresponding to this HIT and automatically approving or rejecting HIT. We discuss one human-in-the-loop task in the second case study.

%% file: sections/case_studies.tex
\section{Case Studies}
\label{sec:case_study}
In this section, we go over two specific past instantiations of challenges organized on our platform to showcase its various capabilities.

\par \noindent
\subsection{Visual Question Answering.} Visual Question Answering (VQA) is a multi-modal task where given an image and a free-form open-ended natural language question about the image, the AI agent's task is to answer the question accurately. The Visual Question Answering Challenge (VQA) 2016 was organized on the 
VQAv1 \cite{{VQA}} dataset
and was hosted on another platform,
where mean evaluation time per dataset instance was $\sim$10 minutes.
In the following years, VQA Challenge 2017 and 2018 (on the VQAv2 \cite{goyal_cvpr17} dataset) were hosted on EvalAI.
Even with twice the dataset size (VQAv2 vs VQAv1), our parallelized implementation offered a significant reduction in per-instance evaluation time ($\sim$130 seconds) -- an approximately 12x speedup. 
This was made possible by leveraging map-reduce techniques to distribute smaller chunks of the test-split on multiple cores and eventually combining the individual results to 
compute overall performance.
Execution time is further reduced by making sure that the evaluation program is not loaded in memory (preloaded earlier) everytime a new submission arrives.
The above instance of the challenge also utilized several other features of our platform -- namely, supporting multiple challenge phases for continued evaluation beyond the challenge; multiple data-splits for debugging submissions and reporting public benchmarks and privacy levels for leaderboards associated with different data-split evaluations.

\begin{figure}[]
\centering
\includegraphics[width=0.80\columnwidth]{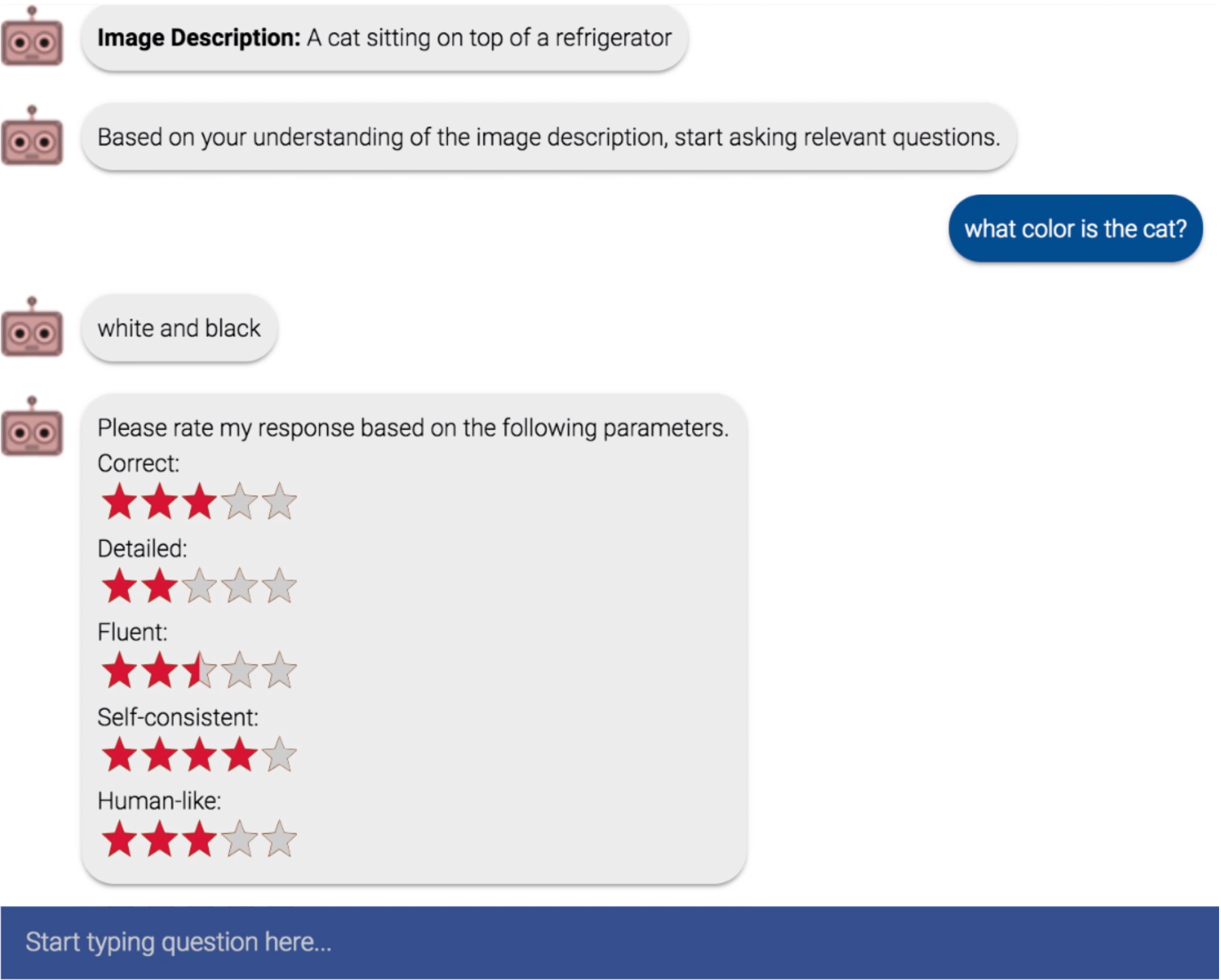}
\caption{Human-in-the-loop interface for evaluating visual dialog agents.}
\label{fig:vis_dial_hieme}
\end{figure}

\par\noindent
\subsection{Visual Dialog.} 
As mentioned before, it is notoriously difficult to evaluate 
free-form multimodal tasks such as image captioning~\cite{dai2017towards,li2018generating}, visual dialog \cite{visdial, visdial_rl,liu2016not} etc using automatic metrics and as such they inherently require human-in-the-loop evaluation. 
Recall that
Visual Dialog, where given an image, an associated dialog history and a follow-up question about the image, an agent is required to answer the question while 
inferring relevant context from
history -- evaluation is further complicated by the huge set of possibly 'correct' answers and the relatively sparse sampling of this space, even in large-scale datasets -- making human-in-the-loop evaluation imperative.
As part of a demonstration at CVPR 2018,
EvalAI hosted a visual dialog challenge where
each submission was connected with a human subject (\figref{fig:vis_dial_hieme}) on Amazon Mechanical Turk tasked with rating a response generated by participating model along several axes -- correctness, fluency, consistency, etc. After 10 such rounds of human-agent interaction, the human's rating of the agent was reflected as a score on the
leaderboard immediately.

\par\noindent
\subsection{Embodied Question Answering.}
Finally, as we move towards developing intelligent agents for tasks situated in \emph{active environments}
instead of \emph{static datasets}, where agents take actions to change the state of the
world around them, it is imperative that we build new tools to
accurately benchmark agents in environments.
One example of such a task is Embodied Question Answering~\cite{embodiedqa} --
an agent is spawned at a random location in a simulated environment (say in a kitchen) and is asked a natural language question (``What color is the car?'').
The agent perceives its environment through first-person vision and can perform a few actions: {\small\texttt{\{move-forward, turn-left, turn-right, stop\}}}.
The agent's objective is to explore the environment and gather visual information necessary to answer the question (``orange'').
Evaluating agents for EmbodiedQA presents a key challenge --
instead of a hidden test dataset, there
are hidden test environments, so participants have to submit pretrained models
and inference code, which has to be reliably executed in these environments to benchmark them.
In ongoing work, we have developed an evaluation framework for EmbodiedQA --
wherein participants upload Docker images with their pretrained models on
Amazon S3, which is then attached and run against test environments
and evaluation metrics provided by the challenge organizer.
We will be using this to host a CVPR 2019 challenge on EmbodiedQA, and aim to
extend this to support a wide range of reinforcement learning task evaluations in future.

\par\noindent
\subsection{fastMRI.} fastMRI~\cite{zbontar2018fastmri} is a collaborative research project between Facebook AI Research (FAIR) and NYU Langone Health to investigate the use of AI to make MRI scans upto 10 times faster. By focusing on the reconstruction capabilities of several AI algorithms, the goal is to enable faster scanning and subsequently making MRIs accessible to more people. This collaborative effort to accelerate Magnetic Resonance Imaging (MRI) by taking fewer measurements was recently structured as a challenge organized around a large-scale open dataset for both raw MR measurements and clinical MR images. EvalAI currently hosts the first iteration of the fastMRI challenge. Ensuring proper benchmarking on such a medical dataset comes with it's own set of challenges -- primarily centered around privacy and storage. Firstly, since proposed algorithms for fastMRI have a direct real world impact, any leakage of test-annotations compromises generalization and can subsequently have drastic consequences. Secondly, in addition to privacy, it is important to note that the dataset itself consumes a lot of storage space as it consists of clinical MR images. As such, supporting decentralized evaluation in addition to a centralized leaderboard to benchmark solutions seems desirable from the organizer's perspective. EvalAI fulfills both of these requirements -- as a platform, we do not have access to the test-annotations for fastMRI but still support efficient evaluation on a remote server (belonging to the organizer). The evaluation metrics are sent from the remote servers to EvalAI via an API format provided by EvalAI.The metrics are then displayed on a  centralized leaderboard hosted on EvalAI.

%% file: sections/conclusion.tex
\section{Conclusion}
\label{sec:conclusion}
While traditional platforms were adequate for evaluation of tasks using automatic metrics, there is a critical need to support human-in-the-loop evaluation for more free-form multimodal tasks like (visual) dialog, question-answering, etc. To this end, we have developed a new evaluation platform that supports the same on a large-scale. Effectively, EvalAI supports pairing an AI agent with thousands of workers so as to rate or evaluate the former over multiple rounds of interaction. By providing a scalable platform that supports such evaluations will eventually encourage the community to benchmark performance on tasks extensively, leading to better understanding of a model's performance both in isolation and in human-AI teams.
\\
\par \noindent 
\textbf{Acknowledgements - }We thank Abhishek Das, Devi Parikh, and Viraj Prabhu for helpful discussions. We would also like to thank Google Summer of Code, Google Code-in, and the 80+ developers and designers who have contributed code to the EvalAI project. This work was supported in part by NSF, AFRL, DARPA, Siemens, Google, Amazon, ONR YIPs and ONR Grants N00014-16-1-{2713,2793}. The views and conclusions contained herein are those of the authors and should not be interpreted as necessarily representing the official policies or endorsements, either expressed or implied, of the U.S. Government, or any sponsor.

%% file: main.bbl
\begin{thebibliography}{10}

\bibitem{amt}
{Amazon Mechanical Turk (AMT)}.
\newblock Website - \url{https://www.mturk.com/}.

\bibitem{VQA}
Stanislaw Antol, Aishwarya Agrawal, Jiasen Lu, Margaret Mitchell, Dhruv Batra,
  C.~Lawrence Zitnick, and Devi Parikh.
\newblock {VQA}: {V}isual {Q}uestion {A}nswering.
\newblock In {\em International Conference on Computer Vision (ICCV)}, 2015.

\bibitem{DBLP:journals/corr/abs-1207-4708}
Marc~G. Bellemare, Yavar Naddaf, Joel Veness, and Michael Bowling.
\newblock The arcade learning environment: An evaluation platform for general
  agents.
\newblock {\em CoRR}, abs/1207.4708, 2012.

\bibitem{Bordes2016LearningEG}
Antoine Bordes and Jason Weston.
\newblock Learning end-to-end goal-oriented dialog.
\newblock {\em CoRR}, abs/1605.07683, 2016.

\bibitem{BrockmanCPSSTZ16}
Greg Brockman, Vicki Cheung, Ludwig Pettersson, Jonas Schneider, John Schulman,
  Jie Tang, and Wojciech Zaremba.
\newblock Openai gym.
\newblock {\em CoRR}, abs/1606.01540, 2016.

\bibitem{visdial_eval}
Prithvijit Chattopadhyay, Deshraj Yadav, Viraj Prabhu, Arjun Chandrasekaran,
  Abhishek Das, Stefan Lee, Dhruv Batra, and Devi Parikh.
\newblock Evaluating visual conversational agents via cooperative human-ai
  games.
\newblock In {\em Proceedings of the Fifth AAAI Conference on Human Computation
  and Crowdsourcing (HCOMP)}, 2017.

\bibitem{CodaLab}
{CodaLab}.
\newblock Website - \url{https://competitions.codalab.org/}.

\bibitem{dai2017towards}
Bo~Dai, Dahua Lin, Raquel Urtasun, and Sanja Fidler.
\newblock Towards diverse and natural image descriptions via a conditional gan.
\newblock {\em 2017 IEEE International Conference on Computer Vision (ICCV)},
  pages 2989--2998, 2017.

\bibitem{embodiedqa}
Abhishek Das, Samyak Datta, Georgia Gkioxari, Stefan Lee, Devi Parikh, and
  Dhruv Batra.
\newblock Embodied question answering.
\newblock {\em 2018 IEEE/CVF Conference on Computer Vision and Pattern
  Recognition}, pages 1--10, 2018.

\bibitem{visdial}
Abhishek Das, Satwik Kottur, Khushi Gupta, Avi Singh, Deshraj Yadav, Jos{\'e}
  M.~F. Moura, Devi Parikh, and Dhruv Batra.
\newblock Visual dialog.
\newblock {\em 2017 IEEE Conference on Computer Vision and Pattern Recognition
  (CVPR)}, pages 1080--1089, 2017.

\bibitem{visdial_rl}
Abhishek Das, Satwik Kottur, Jos{\'e} M.~F. Moura, Stefan Lee, and Dhruv Batra.
\newblock Learning cooperative visual dialog agents with deep reinforcement
  learning.
\newblock {\em 2017 IEEE International Conference on Computer Vision (ICCV)},
  pages 2970--2979, 2017.

\bibitem{django}
{Django: The web framework for perfectionists with deadlines.}
\newblock Website - \url{https://www.djangoproject.com/}.

\bibitem{docker}
{Docker}.
\newblock Website - \url{https://www.docker.com/}.

\bibitem{DBLP:journals/corr/DuanCHSA16}
Yan Duan, Xi~Chen, Rein Houthooft, John Schulman, and Pieter Abbeel.
\newblock Benchmarking deep reinforcement learning for continuous control.
\newblock {\em CoRR}, abs/1604.06778, 2016.

\bibitem{ecs}
{Amazon Elastic Container Service (ECS)}.
\newblock Website - \url{https://aws.amazon.com/ecs/}.

\bibitem{geman2015visual}
Donald Geman, Stuart Geman, Neil Hallonquist, and Laurent Younes.
\newblock Visual turing test for computer vision systems.
\newblock {\em Proceedings of the National Academy of Sciences}, page
  201422953, 2015.

\bibitem{goodfellow_nips14}
Ian~J. Goodfellow, Jean Pouget-Abadie, Mehdi Mirza, Bing Xu, David
  Warde-Farley, Sherjil Ozair, Aaron~C. Courville, and Yoshua Bengio.
\newblock Generative adversarial nets.
\newblock In {\em Advances in Neural Information Processing Systems (NIPS)},
  2014.

\bibitem{goyal_cvpr17}
Yash Goyal, Tejas Khot, Douglas Summers-Stay, Dhruv Batra, and Devi Parikh.
\newblock Making the v in vqa matter: Elevating the role of image understanding
  in visual question answering.
\newblock In {\em Proceedings of IEEE Conference on Computer Vision and Pattern
  Recognition (CVPR)}, 2017.

\bibitem{he2017mask}
Kaiming He, Georgia Gkioxari, Piotr Doll{\'a}r, and Ross Girshick.
\newblock Mask r-cnn.
\newblock In {\em Computer Vision (ICCV), 2017 IEEE International Conference
  on}, pages 2980--2988. IEEE, 2017.

\bibitem{he2016deep}
Kaiming He, Xiangyu Zhang, Shaoqing Ren, and Jian Sun.
\newblock Deep residual learning for image recognition.
\newblock In {\em Proceedings of the IEEE conference on computer vision and
  pattern recognition}, pages 770--778, 2016.

\bibitem{Kaggle}
{Kaggle}.
\newblock Website - \url{https://kaggle.com/}.

\bibitem{kolve2017ai2}
Eric Kolve, Roozbeh Mottaghi, Daniel Gordon, Yuke Zhu, Abhinav Gupta, and Ali
  Farhadi.
\newblock Ai2-thor: An interactive 3d environment for visual ai.
\newblock {\em arXiv preprint arXiv:1712.05474}, 2017.

\bibitem{krizhevsky_nips12}
Alex Krizhevsky, Ilya Sutskever, and Geoff Hinton.
\newblock {ImageNet Classification with Deep Convolutional Neural Networks}.
\newblock In {\em Advances in Neural Information Processing Systems (NIPS)},
  2012.

\bibitem{li2018generating}
Dianqi Li, Xiaodong He, Qiuyuan Huang, Ming-Ting Sun, and Lei Zhang.
\newblock Generating diverse and accurate visual captions by comparative
  adversarial learning.
\newblock {\em arXiv preprint arXiv:1804.00861}, 2018.

\bibitem{lin2014microsoft}
Tsung-Yi Lin, Michael Maire, Serge Belongie, James Hays, Pietro Perona, Deva
  Ramanan, Piotr Doll{\'a}r, and C~Lawrence Zitnick.
\newblock Microsoft coco: Common objects in context.
\newblock In {\em European conference on computer vision}, pages 740--755.
  Springer, 2014.

\bibitem{liu2016not}
Chia-Wei Liu, Ryan Lowe, Iulian~V Serban, Michael Noseworthy, Laurent Charlin,
  and Joelle Pineau.
\newblock How not to evaluate your dialogue system: An empirical study of
  unsupervised evaluation metrics for dialogue response generation.
\newblock {\em arXiv preprint arXiv:1603.08023}, 2016.

\bibitem{long2015fully}
Jonathan Long, Evan Shelhamer, and Trevor Darrell.
\newblock Fully convolutional networks for semantic segmentation.
\newblock In {\em Proceedings of the IEEE conference on computer vision and
  pattern recognition}, pages 3431--3440, 2015.

\bibitem{miller2017parlai}
Alexander~H Miller, Will Feng, Adam Fisch, Jiasen Lu, Dhruv Batra, Antoine
  Bordes, Devi Parikh, and Jason Weston.
\newblock Parlai: A dialog research software platform.
\newblock {\em arXiv preprint arXiv:1705.06476}, 2017.

\bibitem{Rajpurkar2016SQuAD10}
Pranav Rajpurkar, Jian Zhang, Konstantin Lopyrev, and Percy Liang.
\newblock Squad: 100, 000+ questions for machine comprehension of text.
\newblock In {\em EMNLP}, 2016.

\bibitem{redmon2016you}
Joseph Redmon, Santosh Divvala, Ross Girshick, and Ali Farhadi.
\newblock You only look once: Unified, real-time object detection.
\newblock In {\em Proceedings of the IEEE conference on computer vision and
  pattern recognition}, pages 779--788, 2016.

\bibitem{simonyan_iclr15}
Karen Simonyan and Andrew Zisserman.
\newblock {Very deep convolutional networks for large-scale image recognition}.
\newblock In {\em Proceedings of the International Conference on Learning
  Representations (ICLR)}, 2015.

\bibitem{sqs}
{Amazon Simple Queue Service}.
\newblock Website - \url{https://aws.amazon.com/sqs/}.

\bibitem{Tapaswi2016MovieQAUS}
Makarand Tapaswi, Yukun Zhu, Rainer Stiefelhagen, Antonio Torralba, Raquel
  Urtasun, and Sanja Fidler.
\newblock Movieqa: Understanding stories in movies through question-answering.
\newblock {\em 2016 IEEE Conference on Computer Vision and Pattern Recognition
  (CVPR)}, pages 4631--4640, 2016.

\bibitem{OpenML2013}
Joaquin Vanschoren, Jan~N. van Rijn, Bernd Bischl, and Luis Torgo.
\newblock Openml: Networked science in machine learning.
\newblock {\em SIGKDD Explorations}, 15(2):49--60, 2013.

\bibitem{zbontar2018fastmri}
Jure Zbontar, Florian Knoll, Anuroop Sriram, Matthew~J Muckley, Mary Bruno,
  Aaron Defazio, Marc Parente, Krzysztof~J Geras, Joe Katsnelson, Hersh
  Chandarana, et~al.
\newblock fastmri: An open dataset and benchmarks for accelerated mri.
\newblock {\em arXiv preprint arXiv:1811.08839}, 2018.

\bibitem{zhu2017target}
Yuke Zhu, Roozbeh Mottaghi, Eric Kolve, Joseph~J Lim, Abhinav Gupta,
  Li~Fei-Fei, and Ali Farhadi.
\newblock Target-driven visual navigation in indoor scenes using deep
  reinforcement learning.
\newblock In {\em Proceedings of IEEE International Conference on Robotics and
  Automation (ICRA)}, 2017.

\end{thebibliography}
